\documentclass[sigconf]{acmart}
\makeatletter
\def\@ACM@checkaffil{
    \if@ACM@instpresent\else
    \ClassWarningNoLine{\@classname}{No institution present for an affiliation}%
    \fi
    \if@ACM@citypresent\else
    \ClassWarningNoLine{\@classname}{No city present for an affiliation}%
    \fi
    \if@ACM@countrypresent\else
        \ClassWarningNoLine{\@classname}{No country present for an affiliation}%
    \fi
}
\makeatother

\AtBeginDocument{%
  \providecommand\BibTeX{{%
    \normalfont B\kern-0.5em{\scshape i\kern-0.25em b}\kern-0.8em\TeX}}}

\copyrightyear{2023} 
\acmYear{2023} 
\setcopyright{acmlicensed}\acmConference[KDD '23]{Proceedings of the 29th ACM SIGKDD Conference on Knowledge Discovery and Data Mining}{August 6--10, 2023}{Long Beach, CA, USA}
\acmBooktitle{Proceedings of the 29th ACM SIGKDD Conference on Knowledge Discovery and Data Mining (KDD '23), August 6--10, 2023, Long Beach, CA, USA}
\acmPrice{15.00}
\acmDOI{10.1145/3580305.3599296}
\acmISBN{979-8-4007-0103-0/23/08}

\usepackage{multirow}
\usepackage{mathrsfs}
\usepackage{tikz}
\usepackage{cleveref}
\usepackage{balance}

\usepackage{amsmath, color, enumerate, amsthm, graphicx}
\usepackage{caption, subcaption}
\usepackage[ruled,linesnumbered]{algorithm2e}
\usepackage{csvsimple}
\usepackage{enumitem}

\newtheorem{remark}{Remark}

\crefname{section}{Section}{Sections}
\crefname{figure}{Figure}{Figures}
\crefname{theorem}{Theorem}{Theorems}
\crefname{lemma}{Lemma}{Lemmas}
\crefname{remark}{Remark}{Remarks}
\crefname{example}{Example}{Examples}
\crefname{appendix}{Appendix}{Appendicies}
\crefname{myprop}{Proposition}{Propositions}
\crefname{equation}{Eq.}{Equations}
\crefname{table}{Table}{Tables}
\crefname{algorithm}{Algorithm}{Algorithms}
\begin{document}

\newcommand{\camera}[1]{\textcolor{red}{#1}}
\newcommand{\bgm}[1]{\textcolor{red}{Qing: #1}}
\newcommand{\outline}[1]{\textcolor{blue}{outline: #1}}
\newcommand{\rc}[1]{\textcolor{brown}{ruocheng: #1}}
\newcommand{\yl}[1]{\textcolor{blue}{yang: #1}}
\newcommand{\xy}[1]{\textcolor{cyan}{xy: #1}}
\newcommand{\whn}[1]{\textcolor{green}{Wang: #1}}
\newcommand{\jef}[1]{\textcolor{blue}{Jef: #1}}

\newcommand{\va}{\boldsymbol{a}}
\newcommand{\embe}{\textbf{e}}
\newcommand{\embc}{\textbf{c}}
\newcommand{\adc}{\hat{z}}
\newcommand{\adcu}{\hat{z}_u}
\newcommand{\adcub}{\hat{\boldsymbol{z}}_{u}}
\newcommand{\zu}{\boldsymbol{z}_{u} }
\newcommand{\zus}{z_u}
\newcommand{\au}{\boldsymbol{a}_{u}}
\newcommand{\ru}{\boldsymbol{r}_{u}}
\newcommand{\wu}{\boldsymbol{w}_{u}}
\newcommand{\vz}{\boldsymbol{z}}
\newcommand{\vr}{\textbf{r}}
\newcommand{\vw}{\textbf{w}}
\newcommand{\zawsub}{p(\adcub|\au,\wu))}
\newcommand{\rawsub}{p(\Rui|\au, \wu)}
\newcommand{\qphi}{q_\phi(\adcub|\au,\wu)}
\newcommand{\pthetaazw}{p_\theta(A_u=\va,\adcu=z|W_u=w)}
\newcommand{\pthetaaw}{p_\theta(A_u=\va|W_u=w)}
\newcommand{\pthetazw}{p_\theta(\adcu=z|W_u=w)}
\newcommand{\modelname}{iDCF }
\newcommand{\Ra}{r^{\va}}
\newcommand{\Ruia}{r_{ui}^{\va}}
\newcommand{\Rui}{r_{ui}}
\newcommand{\aui}{a_{ui}}

\title{Debiasing Recommendation by Learning Identifiable Latent Confounders}

\author{Qing Zhang}
\authornote{This work was done when the first author was an intern at Bytedance Research.}
\email{qzhangbo@connect.ust.hk}
\affiliation{%
  \institution{Hong Kong University of Science and Technology}
}

\author{Xiaoying Zhang}
\authornote{Corresponding Author.}
\email{zhangxiaoying.xy@bytedance.com}
\affiliation{%
  \institution{ByteDance Research}
}

\author{Yang Liu}
\email{yang.liu01@bytedance.com}
\affiliation{%
  \institution{ByteDance Research}
}

\author{Hongning Wang}
\email{hw5x@virginia.edu}
\affiliation{%
  \institution{University of Virginia}
}

\author{Min Gao}
\email{gaomin@cqu.edu.cn}
\affiliation{%
  \institution{Chongqing Univeristy}
}

\author{Jiheng Zhang}
\email{jiheng@ust.hk}
\affiliation{%
  \institution{Hong Kong University of Science and Technology}
}

\author{Ruocheng Guo}
\email{ruocheng.guo@bytedance.com}
\affiliation{%
  \institution{ByteDance Research}
}

\renewcommand{\shortauthors}{Qing Zhang et al.}

\begin{abstract}

%

%
Recommendation systems aim to predict users' feedback on items not exposed to them yet.
 Confounding bias arises due to the presence of unmeasured variables (e.g., the socio-economic status of a user) that can affect both a user's exposure and feedback. Existing methods either (1) make untenable assumptions about these unmeasured variables or (2) directly infer latent confounders from users' exposure. However, they cannot guarantee the identification of counterfactual feedback, which can lead to biased predictions.
In this work, we propose a novel method, i.e., identifiable deconfounder (iDCF), which leverages a set of proxy variables (e.g., observed user features) to resolve the aforementioned non-identification issue.
The proposed iDCF is a general deconfounded recommendation framework that applies proximal causal inference to infer the unmeasured confounders and identify the counterfactual feedback with theoretical guarantees.  
Extensive experiments on various real-world and synthetic datasets verify the proposed method's effectiveness and robustness.
\end{abstract}





\maketitle

\section{Introduction}
\label{sec: introduction}

Recommendation systems play an essential role in a wide range of real-world applications, such as video streaming \cite{gao2022kuairand}, e-commerce \cite{zhou2018deep}, web search \cite{wu2022survey}.
Such systems aim to expose users to items that align with their preferences by predicting their counterfactual feedback, i.e., the feedback users would give if they were exposed to an item. 
Looking at the recommendation problem from a causal perspective \cite{ROBINS19861393}, a user's counterfactual feedback on an item can be seen as a potential outcome where exposure is the treatment.
However, predicting potential outcomes by estimating the correlation between exposure and feedback can be problematic due to confounding bias. This occurs when unmeasured factors that affect both exposure and feedback, such as the socio-economic status of the user, are not accounted for.
 For example, on an e-commerce website, users of higher socio-economic status are more likely to be exposed to expensive items because of their history of higher-priced consumption.
 These users might also tend to give negative feedback for products due to their higher standards for item quality.
 Without proper adjustment for this unmeasured confounder, recommendation models may pick up the spurious correlation that expensive items are more likely to receive negative feedback.
As a result, it is essential to mitigate the confounding bias to guarantee the identification of the counterfactual feedback
which is a prerequisite for the accurate prediction of a user's feedback through data-driven models in recommender systems. 

Recent literature has proposed various methods to address the issue of confounding bias in recommender systems. These methods can be broadly split into two settings: measured confounders and unmeasured confounders.
%
For the measured confounders (e.g., item popularity and video duration) that can be obtained from the dataset,
previous work applies standard causal inference methods such as backdoor adjustment \cite{reason:Pearl09a} and inverse propensity reweighting \cite{IPS-16} to mitigate the specific biases \cite{backdoor-group-pop, backdoor-item-pop, backdoor-video}.

In practice, it is more common for there to be unmeasured confounders that cannot be accessed from the recommendation datasets due to various reasons, such as privacy concerns,
e.g., users usually prefer to keep their socio-economic statuses private from the system. 
Alternatively, in most real-world recommendation scenarios, one even does not know what or how many confounders exist. 
%
In general, it is impossible to obtain an unbiased estimate of the potential outcome, i.e., the user's counterfactual feedback, without additional related information about unmeasured confounders \cite{kuroki2014measurement}. 

\begin{figure}[t]
    \centering
    \includegraphics[scale=0.1]{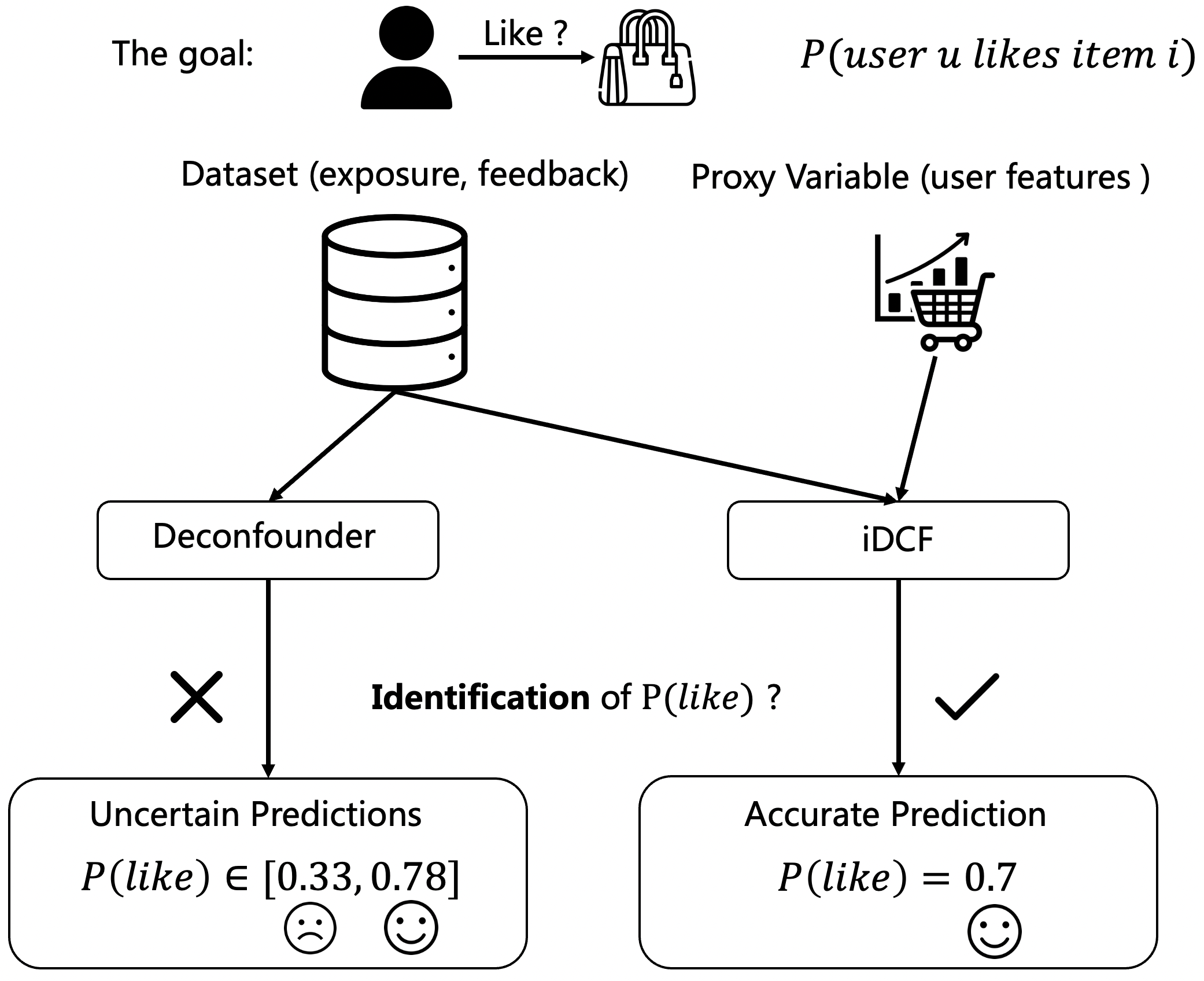}
    \caption{
When predicting the user's feedback, 
    the non-identification of the user's counterfactual feedback will make the recommendation method yield different feasible predictions (probabilities in the interval) that are compatible with the given dataset and will not converge, even with infinite data,
    leading to the uncertainty of the user's feedback.
    See \cref{example: counterexample} for more details.
}
    \label{fig: non-identification}
\vspace{-6mm}
\end{figure}

As a result, previous methods have relied on additional assumptions regarding unmeasured confounders. 
For example, RD-IPS \cite{address} assumes the bounded impact of unmeasured confounders on item exposure and performs robust optimization for deconfounding. 
Invariant Preference Learning \cite{wang2022invariant} relies on the assumption of several abstract environments as the proxy of unmeasured confounders and applies invariant learning for debiasing.
However, these methods heavily rely on assumptions about unmeasured confounders and do not provide a theoretical guarantee of the identification of the potential outcome \cite{reason:Pearl09a} .  
Another line of methods, such as \cite{IV-serach-data,frontdoor-dcf,front-door-cvr}, assume the availability of an additional instrumental variable (IV), such as search log data, or mediator, such as click feedback to perform classical causal inference, such as IV-estimation and front door adjustment \cite{reason:Pearl09a}. 
However, it is hard to find and collect convincing instrumental variables or mediators that satisfy the front door criteria \cite{hernan2010causal, reason:Pearl09a} from recommendation data.
Different from previous methods, 
Deconfounder \cite{deconfounder} does not require additional assisted variables and approximates the unmeasured confounder with a substitute confounder learned from the user's historical exposure records.
Nevertheless, it has the inherent non-identification issue \cite{d2019multi,grimmer2020ive},
which means Deconfounder cannot yield a unique prediction of the user's feedback given a fixed dataset.
\cref{fig: non-identification} shows such an example where the recommender model yields different feasible predictions of users' feedback due to the non-identification issue. 

Hence, a glaring issue in the current practice in the recommender systems falls onto the identifiability of the user's counterfactual feedback (potential outcome) in the presence of unmeasured confounders.
This paper focuses on identifying the potential outcome by mitigating the unmeasured confounding bias.


As the user's exposure history is helpful but not enough to infer the unmeasured confounder and identify the counterfactual feedback,
additional information is required.  
Fortunately, such information can be potentially accessed through users' features and historical interactions with the system.  
Using the previous example,  while the user's socio-economic status (unmeasured confounder) cannot be directly accessed, 
we can access the user's consumption level from his recently purchased items, 
whose prices will be beneficial in inferring the user's socio-economic status.

To this end, we formulate the debiasing recommendation problem as a causal inference problem with multiple treatments (different items to be recommended), 
and utilize the proximal causal inference technique \cite{proximal-introduction} 
which  assumes the availability of a proxy variable (e.g., user's consumption level), 
which is a descendant of the unmeasured confounder (e.g., user's socio-economic status).
Theoretically, the proxy variable can help infer the unmeasured confounder and the effects of exposure and confounders on the feedback.
This leads to the identification of the potential outcome (see our \cref{thm: identification}), which is crucial for accurate predictions of users' counterfactual feedback to items that have not been exposed.
Practically, we choose user features as proxy variables since they are commonly found in recommender system datasets 
and the theoretical requirement of proxy variables is easier to be satisfied compared with the instrumental variables and mediators 
\cite{miao2022identifying}.

Specifically, we propose a novel approach to address unmeasured confounding bias in the task of debiasing recommender system, referred to as the \emph{identifiable deconfounder} (iDCF). The proposed method is feedback-model-agnostic and can effectively handle situations where unmeasured confounders are present.
%
\modelname utilizes the user's historical interactions and additional observable proxy variables to infer the latent confounder effectively with identifiability.
Then, the learned confounder is used to train the feedback prediction model that estimates the combined effect of confounders and exposure on the user's feedback.
In the inference stage, the adjustment method \cite{ROBINS19861393} is applied  to mitigate confounding bias by taking the expectation over the learned confounder.

%
We evaluate the effectiveness of \modelname on a variety of datasets, including both real-world and synthetic, 
which demonstrate its encouraging performance and robustness regarding different confounding effects and data density in predicting user feedback. 
Moreover, on the synthetic dataset with the ground-truth of the unmeasured confounder known, 
we also explicitly show that \modelname can learn a better latent confounder in terms of identifiability.


Our main contributions are summarized as follows:
\begin{itemize}[noitemsep,topsep=0pt,leftmargin=*]
    \item We highlight the importance of identification of potential outcome distribution in the task of debiasing recommendation systems. 
    Moreover, we demonstrate the non-identification issue of the Deconfounder method, which can lead to inaccurate feedback prediction due to confounding bias.
    \item We propose a general recommendation framework that utilizes proximal causal inference to address the non-identification issue in the task of debiasing recommendation systems and provides theoretical guarantees for mitigating the bias caused by unmeasured confounders.
    \item We conduct extensive experiments to show the superiority and robustness of our methods in the presence of unmeasured confounders.
\end{itemize}

\section{Related work}
\subsection{Deconfounding in Recommendation}
As causal inference becomes a popular approach in debiasing recommendation systems and examining relationships between variables \cite{ROBINS19861393, reason:Pearl09a}, researchers now focus more on the challenge of confounding bias.
 Confounding bias is prevalent in recommendation systems due to various confounding factors.
For example, item popularity can create a popularity bias and be considered as a confounder. Several studies have addressed specific confounding biases, such as item popularity \cite{backdoor-item-pop,backdoor-group-pop, wei2021model}, video duration \cite{backdoor-video}, video creator \cite{backdoor-video-creator}, and selection bias \cite{selection-bias}.

However, many unmeasured confounders may also exist, which make the classical deconfounding methods like inverse propensity weighting (IPW) not applicable.
To deal with the confounding bias in the presence of unmeasured confounders, \cite{address} assumes a bounded confounding effect on the exposure and applies robust optimization to improve the worst-case performance of recommendation models, \cite{front-door-cvr, frontdoor-dcf, IV-serach-data} take additional signals as mediators or instrumental variables to eliminate confounding bias. 
\cite{wang2022invariant} assumes the existence of several environments to apply invariant learning.
As shown in our later experiments, these additional strong assumptions on unmeasured confounders can lead to sub-optimal recommendation performance. Moreover, they also fail to provide a theoretical guarantee of the identification of users' counterfactual feedback. 

There is also another line of work \cite{deconfounder, zhu2022deep} that considers the multiple-treatment settings \cite{wang2019blessings} and infers substitute confounders from the user's exposure to incorporate them into the preference prediction models.
However, these methods cannot guarantee the identification of the user's preference,
which may lead to inconsistent, thus poor recommendation performance.

\subsection{Proximal Causal Inference}
Proximal causal inference \cite{kuroki2014measurement,proximal-introduction, proximal-miao-2018, proximal-miao-bridge} assumes the existence of proxy variables of unmeasured confounders in the single-treatment regime,
and the goal is to leverage proxy variables to identify causal effects.
Kuroki and Pearl \cite{kuroki2014measurement} study the identification strategy in the different causal graphs. 
Miao et al. \cite{proximal-miao-2018} generalize their strategy and show nonparametric identification of the causal effect with two independent proxy variables.
Miao et al. \cite{proximal-miao-bridge} further use negative control exposure/outcome to explain the usages of proxy variables intuitively.
However, these methods usually rely on informative proxy variables to infer the unmeasured confounders,
while our method formulates the recommendation problem in the multiple treatment setting,
which enables us to leverage information from the user's exposure to infer the unmeasured confounder. This 
relaxes the requirement on the proxy variables and still theoretically guarantees the identification of the potential outcome \cite{miao2022identifying}.



\section{Problem formulation}
In this section, we first analyze the recommendation problem from a causal view in the presence of unmeasured confounders.
Then we show that Deconfounder \cite{deconfounder}, one of the widely-used methods for recommendations with unobserved confounder, suffers the non-identification issue, i.e., it cannot predict the user's preference consistently, through an illustrative example. This observation motivates our method, which we will detail in the next section. 


\subsection{Notations}
\label{sec: 3.1}
We start with the notations used in this work.
Let scalars and vectors be signified by lowercase letters (e.g., $a$) and boldface lowercase letters (e.g., $\va$), respectively.
Subscripts signify element indexes. For example, $a_i$ is the $i$-th element of the vector $\va$.
The superscript of a  potential outcome denotes its corresponding treatment (e.g., $\Ruia$).

\begin{figure}[t]
\center
\begin{subfigure}{0.5 \textwidth}
\begin{tikzpicture}[->,shorten >=1pt,auto,node distance=4cm,-stealth,
                thick,base node/.style={circle,draw,minimum size=48pt}, real node/.style={double,circle,draw,minimum size=50pt},scale = 1.5]
        \node[shape=circle,draw=black, scale=1.8](1) at(0.2,0.5) { };
        \node[ ] (0) at(1.3,0.5){Observed Variable};
        \node[shape=circle, draw=black, fill=gray, scale=1.8 ] (0) at(3,0.5){};
        \node[ ] (0) at(4.2,0.5){Unmeasured Variable};
        
\end{tikzpicture}

\label{fig: naive graph}
\end{subfigure}

\begin{subfigure}{0.23 \textwidth}

 \begin{tikzpicture}[->,shorten >=1pt,auto,node distance=4cm,-stealth,
                thick,base node/.style={circle,draw,minimum size=48pt}, real node/.style={double,circle,draw,minimum size=50pt},scale = 1.5]
        \node[shape=circle, draw=black, fill=gray,scale=0.9 ] (0) at(1,0.9){$\zu$};
        \node[shape=circle,draw=black,scale=0.8](1) at(0,0) {$a_{u1}$};
        \node[shape=circle,draw=black,scale=0.8](2) at(0.85,0) {$a_{u2}$};
        \node[](3) at(1.35,0) {$ \cdot \cdot \cdot $};
        \node[shape=circle,draw=black,scale=0.8](4) at(2.3,0) {$a_{un}$};
        
        \node[shape=circle,draw=black,scale=0.8](5) at(0,-1) {$r_{u1}$};
        \node[shape=circle,draw=black,scale=0.8](6) at(1.3,-1) {$r_{u2}$};
        \node[](8) at(1.8,-1) {$ \cdot \cdot \cdot $};
        \node[shape=circle,draw=black,scale=0.8](7) at(2.3,-1) {$r_{un}$};

        \path[]
        (0) edge node {} (1)
        (0) edge node {} (2)
        (0) edge node {} (4)
        (1) edge node {} (5)
        (2) edge node {} (6)
        (4) edge node {} (7)
        (0) edge node {} (5)
        (0) edge node {} (6)
        (0) edge node {} (7)
        ;
\end{tikzpicture}
\caption{Without proxy variables}    

\end{subfigure}
\begin{subfigure}{0.23\textwidth}
 \begin{tikzpicture}[->,shorten >=1pt,auto,node distance=4cm,-stealth,
                thick,base node/.style={circle,draw,minimum size=48pt}, real node/.style={double,circle,draw,minimum size=50pt},scale = 1.5]
        \node[shape=circle, draw=black, fill=gray,scale=0.9 ] (0) at(1,0.9){$\zu$};
        \node[shape=circle,draw=black,scale=0.8](8) at(-0.1,0.9) {$\wu$};
        \node[shape=circle,draw=black,scale=0.8](1) at(0,0) {$a_{u1}$};
        \node[shape=circle,draw=black,scale=0.8](2) at(0.85,0) {$a_{u2}$};
        \node[](3) at(1.35,0) {$ \cdot \cdot \cdot $};
        \node[shape=circle,draw=black,scale=0.8](4) at(2.3,0) {$a_{un}$};
        
        \node[shape=circle,draw=black,scale=0.8](5) at(0,-1) {$r_{u1}$};
        \node[shape=circle,draw=black,scale=0.8](6) at(1.3,-1) {$r_{u2}$};
        \node[](9) at(1.8,-1) {$ \cdot \cdot \cdot $};
        \node[shape=circle,draw=black,scale=0.8](7) at(2.3,-1) {$r_{un}$};

        \path[]
        (0) edge node {} (1)
        (0) edge node {} (2)
        (0) edge node {} (4)
        (1) edge node {} (5)
        (2) edge node {} (6)
        (4) edge node {} (7)
        (0) edge node {} (5)
        (0) edge node {} (6)
        (0) edge node {} (7)
        (0) edge node {} (8)
        ;
\end{tikzpicture}
\caption{With a proxy variable}
\label{fig: proximal causal graph}
\end{subfigure}
\caption{Causal graphs with multiple treatments for recommendation systems.
 $\va_{ui}$: exposure of user $u$ to item $i$, 
 $\Rui$: feedback of user $u$ on item $i$,
 $\zu$: the unmeasured confounder,
 $\wu$: a proxy variable of the confounder.
 }

\label{fig: two causal graph}
\vspace{-4mm}
\end{figure}
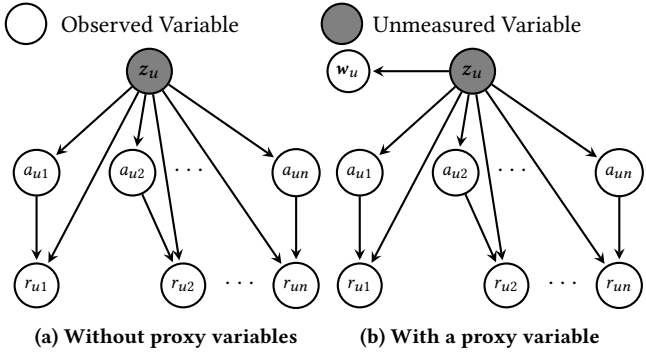

We adopt the potential outcome framework \cite{rubin1974estimating} with multiple treatments \cite{wang2019blessings} to formulate the problem.
The causal graph is shown in \cref{fig: two causal graph}.
Let $\mathcal{U}=\{u\}$ and $\mathcal{I}=\{i\}$ denote the set of users and items, respectively with $|\mathcal{U}|=m, |\mathcal{I}|=n$.
We define the following components of the framework:
\begin{itemize}[noitemsep,topsep=0pt,leftmargin=*]
    \item Multiple treatments: $\au = [a_{u1}, a_{u2}, \dots, a_{un} ]\in \{0,1\}^n$ is the observed exposure status of user $u$, where $a_{ui} = 1$ ($a_{ui}=0$) means item $i$ was exposed to user $u$ (not exposed to user $u$) in history.

   \item Observed outcome: $r_{ui}$ denotes the observed feedback of the user-item pair $(u,i)$ and $\ru = [r_{u1},...,r_{un}]$ signifies the observed feedbacks of user $u$. 
       
    \item 
    Potential outcome:
    %
    $\Ruia $ denotes the potential outcome    \footnote{The distribution of potential outcome $\Ruia$ is equivalent to $p(r_{ui}|do(\va))$ in the structural causal model (SCM) framework.} that would be observed if the user's exposure had been set to the vector value $\va \in \{0,1\}^n$.
    Following previous work~\cite{deconfounder}, we assume $\Ruia$ is only affected by the exposure of item $i$ to user $u$. 

    \item 
    Unmeasured confounder: $\zu \in \mathbb{R}^{d} $  (e.g., the user's socio-economic status) denotes the d-dimensional unmeasured confounder that causally influences both user's exposures $\au$ and feedback $\ru$. 

\end{itemize}

\noindent\textbf{Problem Statement.} Given observational data $\{\au,\ru\}_{u \in \mathcal{U}}$, a recommendation algorithm aims to accurately predict the feedback of user $u$ on item $i$ if the item had been exposed to $u$, i.e., the expectation of the potential outcome $E[\Ruia]$,
where $a_i=1$.
Practically, for a user $u$, items are ranked by the predicted $E[\Ruia]$ such that the user will likely give positive feedback to items ranked in top positions.

However, in real-world scenarios, as the data of the recommendation system is naturally collected as users interact with the recommended items without randomized controlled trials, there usually exists some confounder, $\zu$ as shown in \cref{fig: two causal graph}, which affects both the user $u$'s exposure status $\au$ (i.e., the treatment) and the user's feedback on items $\ru$ (i.e., the outcome),  resulting in possible spurious correlations 
when the user's feedback is simply estimated by $p(\Rui|\au)$.
For instance, the user’s socio-economic status can lead to confounding bias when predicting the user’s counterfactual feedback (See the example in \cref{sec: introduction}). 

Previous work \cite{backdoor-group-pop,backdoor-item-pop,backdoor-video,aspire} takes $\zu$ as a specific factor, for example, item popularity, video duration, video creators, etc. 
But under most real-world circumstances, we cannot access the complete information of $\zu$. 
Thus, this work focuses on a more general problem setting where $\zu$ is an unmeasured confounder.
As shown in \cref{fig: naive graph}, $\zu$ is a latent variable represented by a shaded node.




\subsection{Identification with unmeasured confounder}
\label{sec: 3.2}
To learn the counterfactual feedback from user $u$ on item $i$, i.e., $E[\Ruia]$, 
the identification of the potential outcome distribution $p(\Ruia)$ from observational data is required.
In general, accurately predicting a user's feedback through data-driven models is only possible when causal identifiability has been established.

When all confounders are \textit{measured}, 
 $p(\Ruia)$ can be identified through the classical g-formula\footnote{g-formula is equivalent to the backdoor adjustment in the SCM framework.} \cite{ROBINS19861393}  as follows: 
\begin{equation}
\label{backdoor}
    p(\Ruia) = E_{\zu}[p(\Rui|\va, \zu)].
\end{equation}
When an unmeasured confounder exists, as shown in \cref{fig: naive graph}, it becomes much more challenging to identify $p(\Ruia)$ as the g-formula is no longer applicable.
Previously, Wang et al. \cite{deconfounder} assumed the unmeasured confounder $\zu$ is a common cause of each exposure $a_{ui}$
and proposed  Deconfounder to learn $p(\Ruia)$ with unmeasured confounders.
Deconfounder first learns a substitute confounder $\adcub$ from the exposure vector $\au$ to approximate the true confounder $\zu$,  and directly applies the g-formula in \cref{backdoor} to learn  $p(\Ruia)$.

\noindent
\textbf{Non-Identification of Deconfounder.}
While the high-level idea is promising, Deconfounder fails to guarantee the identification of   $p(\Ruia)$ \cite{d2019multi,grimmer2020ive}.
As shown by the following example, even in a relatively optimistic case where the substitute confounder $\adcu$ can be uniquely determined from the exposure vector $\au$, 
 Deconfounder cannot identify $p(\Ruia)$.
 In other words, $p(\Ruia)$ takes different values under different circumstances,  leading to the inconsistent prediction of the user's feedback.





\begin{example}[Failure of Deconfounder \cite{deconfounder} in  identification.]
\label{example: counterexample}
Consider a recommendation scenario following the causal graph in Figure \ref{fig: two causal graph}, with the confounder $\zus$, the exposure status $\aui$, and the feedback $\Rui$ assumed to be binary random variables. 


We assume $n \geq 3$ to ensure a unique factorization of $p(\au)$ \cite{10.5555/120565.120567}, such that the inferred substitute confounder $\adcu$ in Deconfounder can be uniquely identified from exposure vector $\va$.
In other words, $p(\adcu=1)$ and $p(\adcu=1 |\va)$ are known.
Besides, $p(\Rui=1|\va)$, the probability that user $u$ will give positive feedback to the item $i$ condition on exposure vector $\va$, can also be inferred given a dataset.

Recall that Deconfounder learns $p(\Ruia)$ by applying the g-formula in \cref{backdoor} with the inferred  substitute confounder $\adcu$ as follows:
\begin{equation}
\label{eq: example_identification}
\begin{array}{lcl}
    p(\Ruia=1)  & = & \sum_{z\in \{0,1\}} p(\adcu=z)p(\Ruia=1|\va,\adcu=z) \\
    & = & \sum_{z\in \{0,1\}} p(\adcu=z) \frac{p(\Ruia=1,\adcu=z|\va) }{p(\adcu=z|\va)}.\\
\end{array}
\end{equation} 
For ease of illustration, we denote 
$p_{01|\va} :=p(\adcu=0, \Rui=1|\va), 
\pi_{\adcu=1}:=p(\adcu=1),
\pi_{\adcu=1|\va}:=p(\adcu=1|\va),
\pi_{\Rui=1|\va}:=p(\Rui=1|\va)
$ 
in the rest of the paper, then we get: 
\begin{equation}
\label{eq: cal Rui}
    p(\Ruia=1)=(1-\pi_{\adcu=1}) \frac{p_{01|\va}}{1-\pi_{\adcu=1|\va}} + \pi_{\adcu=1} \frac{p_{11|\va}}{\pi_{\adcu=1|\va}}
\end{equation}

As we assumed before,  $\pi_{\adcu=1}$ and $\pi_{\adcu=1|\va}$ are known, thus it remains to identify $p_{01|a}$, and $p_{11|a}$ to calculate $p(\Ruia=1)$. 
However,  $p_{01|a}$ and $p_{11|a}$ can not be uniquely determined since there are four unknown entries $\{p_{zr|a}, z,r \in\{0,1\}\}$ with three constraints: 
\begin{equation}
\label{eq: constraints}
\sum_{z } \sum_{r} p_{zr|a}=1, 
\sum_{r} p_{1r|a}=\pi_{\adcu=1|a}, 
\sum_z p_{z1}=\pi_{\Rui=1|a},
\end{equation}
\noindent
where the first constraint is the normalization of joint probabilities, and the next two are marginal constraints.
For example, the second constraint is due to 
\begin{equation}
p(\adcu=1,\Rui=0|\va)+p(\adcu=1,\Rui=1|\va)=p(\adcu=1|\va).
\end{equation}
When $\pi_{\adcu=1|\va}$ is not degenerated, i.e.,$\pi_{\adcu=1|\va} \neq 1$ or $0$, which  holds in the recommendation scenario,  the four unknown entries cannot be uniquely determined because there remains one degree of freedom \cite{strang1993introduction}. 
For example, if we take $p_{11|a}$ as the free variable, then it can be any value in the following feasible range:
$$
\max\{0,\pi_{\adcu=1|\va}+\pi_{\Rui=1|\va}-1\} \leq p_{11|a} \leq \min\{\pi_{\adcu=1|\va}, \pi_{\Rui=1|\va}\},
$$
implying $p(\Ruia=1)$ calculated as in \cref{eq: cal Rui} will also be in a range. 
In other words,
\textit{$p(\Ruia=1)$ cannot be identified. }

To make it more explicit, consider this concrete example. 
Assuming $\pi_{\adcu=1}=0.5, \pi_{\adcu=1|\va}=0.2, \pi_{\Rui=1|\va}=0.6$, then $p_{11|\va}$ can be any value in  $[0, 0.1]$,
leading to the feasible range of $p(\Ruia) \in [0.33, 0.78]$.
When the commonly used prediction threshold of $0.5$ is applied, one will get an inconsistent prediction of the user $u$'s preference over item $i$,  since   $p(\Ruia=1)=0.78$ implies user $u$ will prefer the item $i$, while $p(\Ruia=1)=0.33$  not. 
Obviously, the distortion will be even larger when directly ranking items according to  $p(\Ruia=1)$ when its  identification cannot be guaranteed. 
\end{example}

\section{Method}

In this section, we show how proximal causal inference \cite{proximal-introduction} can help ensure the identification of user's counterfactual feedback $p(\Ruia)$.
We start by showing that with a proxy variable (e.g., user features), one can identify $p(\Ruia)$ in \cref{example: counterexample}.
We then propose a feedback-model-agnostic framework iDCF, for the identification of user's counterfactual feedback with unmeasured confounders in general recommendation scenarios with a theoretical guarantee. 

\subsection{Framework}
\label{sec:framework}
A natural question following \cref{example: counterexample} is: 
How to fix the identification issue of  $p(\Ruia)$ so as to predict users' counterfactual feedback uniquely and accurately?
Intuitively, if more information about the unmeasured confounder can be provided, i.e., more constraints in \cref{eq: constraints}, 
$p(\Ruia)$ can be uniquely determined,
which motivates the usage of proximal causal inference \cite{proximal-introduction}.


Inspired by the above intuition, we reformulate the recommendation problem with the unmeasured confounder from the view of proximal causal inference. 
Specifically, we assume that one can observe additional information of user $u$, called proxy variable $w_u$, 
which is directly affected by the unmeasured confounder $\zu$ and 
independent of the feedback $\Rui$ given the unmeasured confounder $\zu$ and the exposure vector $\au$, i.e.,:
$$
\wu=g(\zu), \wu \perp \Rui | (\zu, \au),
$$
where $g$ is an unknown function.
Fortunately, in the recommendation scenario,
such a proxy variable can be potentially accessed through the user's features, 
including user profiles summarized from interaction history. 
For example, when the unmeasured confounder $\zu$ is the user's socio-economic status,
which usually cannot be directly accessed, possibly due to privacy concerns, 
one can take the proxy variable as the average price of items that the user recently purchased,
which is pretty helpful in inferring the user's socio-economic status since high consumption often implies high socio-economic status.
Moreover, such a proxy variable will not directly affect the user's feedback if the user's socio-economic status is already given.



We first show that the user's counterfactual feedback $p(\Ruia)$ in \cref{example: counterexample} can be identified
with a proxy variable $\wu$.

\begin{example}[Success in identifying $p(\Ruia)$ with proxy variable]
\label{example:fix}

Following the settings in \cref{example: counterexample},
we introduce an observable proxy variable $w_u$ that indicates the user's consumption level  affected by the socio-economics status $z_u$ in the recommendation platform, 
the corresponding causal graph is shown in \cref{fig: proximal causal graph}.
We assume $w_u$ is a Bernoulli random variable with mean $\mu(\zu) \in (0,1)$ and $w_u$ is correlated with $z_u$ condition on $\au$.

Similar to \cref{example: counterexample},
$p(\Rui=1|\va, w_u)$, 
the probability that user $u$ will give positive feedback to item $i$ with given exposure status $\va$ and consumption level $w_u$, can be inferred from the given dataset,
and $p(\adcu|\va,w_u)$ is assumed to be uniquely determined by factor models.
Again, for the ease of illustration, we denote  $\pi_{\Rui=1|\va,w} := p(\Rui=1|\au=\va,w_u=w)$ and $\pi_{\adcu=1|\va,w}:=p(\adcu=1|\au=\va,w_u=w)$.
Now, while there are still four unknown entries $\{p_{zr|a}, z,r \in\{0,1\}\}$ as in \cref{eq: constraints}, 
the number of constraints increases from three to four with the two conditional marginal distributions
 $\pi_{\Rui=1|\va,w=0}$ and $\pi_{\Rui|\va,w=1}$, i.e., 
\begin{equation}
        \label{eq: equation set}
        \begin{array}{ll}
                \sum_{z } \sum_{r} p_{zr|\va}=1, & 
                \sum_{z} p_{z1|\va} \frac{\pi_{\adcu=z|\va ,w=1}}{\pi_{\adcu=z|\va}}=\pi_{\Rui=1|\va,w=1}, \\
                \sum_{r} p_{1r|\va}=\pi_{\adcu=1|\va},  
                & 
                \sum_{z} p_{z1|a}  \frac{\pi_{\adcu=z|\va,w=0}}{\pi_{\adcu=z|\va}}=\pi_{\Rui=1|\va,w=0}.
        \end{array}
\end{equation}

The following lemma shows 
the identification result of $p(\Ruia)$.
\begin{lemma}
\label{lemma: identification in example}
There exists a unique solution of   $\{p_{zr|a}, z,r \in\{0,1\}\}$ from \cref{eq: equation set}, leading to the identification of the potential outcome  $p(\Ruia)$ calculated from  \cref{eq: example_identification}.
\end{lemma}

\end{example}


\noindent
\textbf{General framework of identifying the user's counterfactual feedback $ p(\Ruia)$ with proxy variables.}
Next, we show how to identify $ p(\Ruia)$ with proxy variables in general.
Observing that
\begin{equation}
\label{eq: cal po}
\begin{split}
    p(\Ruia) &= E_{\adcub}[p(\Rui|\va,\adcub)] 
              = \int_{\vz} p(\adcub=\vz)p(\Rui|\va,\adcub=\vz) d \vz,
\end{split}
\end{equation}
the key is to infer $p(\adcub)$ and $p(\Rui|\va,\adcub)$,
 yielding the following two-step procedure of the proposed method iDCF: 

\begin{itemize}[noitemsep,topsep=0pt,leftmargin=*]
        \item \textbf{Learning Latent Confounder:} 
        This stage aims to learn a latent confounder $\adcub$ with the help of proxy variables, 
        such that the learned $\adcub$ is equivalent to the true unmeasured confounder $\zu$ up to some transformations \cite{khemakhem2020variational,miao2022identifying}
        and can provide additional constraints to infer the user's feedback $\Rui$, 
        which cannot be achieved by the substitute confounder in Deconfounder. 
        Specifically, we aim to learn its prior distribution, i.e., $ p(\adcu)$.
        Since 
        \begin{equation}
                p(\adcub=z) =E_{\au,\wu}[p(\adcub|\au,\wu)],
            \label{equ:latent-confounder}        
            \end{equation}
        and $p(\au,\wu)$ is measured from the dataset,
        thus the main challenge is to learn $p(\adcu|\au,\wu)$, which can be learned by reconstructing the exposure vector $\au$ based solely on $\wu$, since: 
       \begin{equation}
        \label{eq: factor model with w}
       p(\au|\wu)=\int_{\vz} p(\au|\adcub=\vz)p(\adcub=\vz|\au,\wu)d \vz.
       \end{equation}
       For example, we can apply the widely-used iVAE \cite{khemakhem2020variational} model,
       then $p(\adcub|\au,\wu)$ and  $p(\au|\adcub)$ are estimated by the encoder and the decoder respectively.
     \item \textbf{Feedback with given latent confounder:}
     This stage aims to learn  $p(\Rui|\au,\adcub)$, i.e., user $u$'s feedback on item $i$ 
     under the fixed exposure vector $\au$ and  latent confounder $\adcub$.
     With the help of $p(\adcub|\au,\wu)$ learned in the first stage, one can infer it by 
     directly fitting the observed users' feedback $\rawsub$, since: 
   \begin{equation}
   \label{eq: cal cp}
       \rawsub = \int_z p(\Rui|\adcu=z,\au) p(\adcu=z|\au,\wu)dz.
   \end{equation}

\end{itemize}

Then the potential outcome (i.e., the user's counterfactual feedback) distribution $p(\Ruia)$ is identified by applying \cref{eq: cal po}.
The following theorem shows the general theoretical guarantee
 on identification of $ p(\Ruia)$ through the aforementioned two-step procedure. 
 \begin{figure}
    \centering
    \includegraphics[scale=0.09]{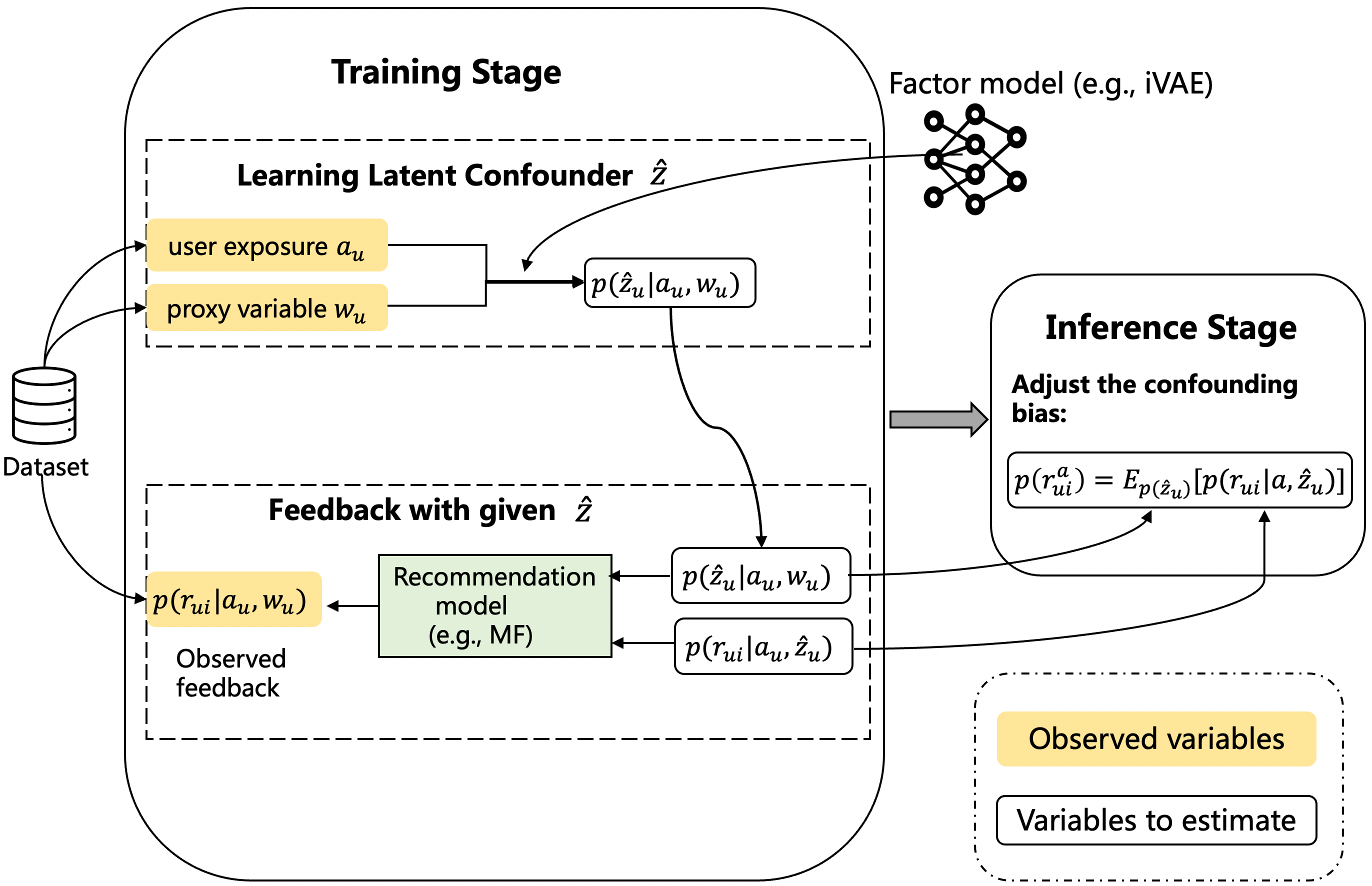}
    \caption{The framework of the proposed method iDCF.}
    \label{fig: framework}
 \vspace{-6mm}
\end{figure}

\begin{theorem}[Identification with proxy variable \cite{miao2022identifying}]
\label{thm: identification}
Under the consistency, ignorability, positivity, exclusion restriction, equivalence, and completeness assumptions,
for any latent joint distribution $p(\au,\adcub|\wu)$ that solves $p(\au|\wu)=\int_{\vz}p(\au,\adcub=\vz|\wu)d \vz$,
there exists a unique solution $p(\Rui|\adcub,\au)$ to the equation \cref{eq: cal cp}
and the potential outcome distribution is identified by \cref{eq: cal po}.
\end{theorem}

\begin{remark}[About assumptions]
Note that \cref{thm: identification} relies on several assumptions:
consistency, ignorability, positivity, exclusion restriction, equivalence, and completeness.
The first 3 assumptions are standard assumptions in causal inference \cite{backdoor-group-pop, backdoor-video}.
Informally, exclusion restriction requires the proxy variable to be independent of the user's feedback conditioned on the confounder and exposure,
which can be reasonable in a recommendation system since the proxy variable (e.g., user's consumption level) is mainly used to implicitly infer the hidden confounder (e.g., user's income) that directly affects user's feedback.
Equivalence requires the unmeasured confounder can be identified from the dataset up to a one-to-one transformation, 
which is also feasible with various factor models \cite{khemakhem2020variational, 10.5555/120565.120567}.
Completeness requires that the proxy variable contains enough information to guarantee the uniqueness of the statistic about the hidden confounder,
which can also be feasible in recommendation scenarios
since the variability in the unmeasured confounders (e.g., user's socio-economics status) is usually captured by variability in the user features (e.g., user's consumption level).

\end{remark}


\subsection{Practical Implementation}
Next, we describe how the proposed iDCF implements the identification steps described in Section~\ref{sec:framework} practically. 
We need to specify:
\begin{itemize}[noitemsep,topsep=0pt,leftmargin=*]
    \item \textbf{Training Stage:} (1) How to learn the latent confounder, i.e.,  $ p(\adcub)$ in \cref{equ:latent-confounder}? As discussed in Section~\ref{sec:framework}, the main challenge is to learn $p(\adcu|\au,\wu)$.
    (2) How does our method learn users' feedback with the given latent confounder, i.e., $p(\Rui|\au,\adcub)$?

   \item \textbf{Inference Stage.} With $ p(\adcub)$ and $p(\Rui|\au,\adcub)$ learned in the training stage, how does the proposed iDCF framework infer the unbiased feedback of users following  \cref{eq: cal po}? 
\end{itemize}

\noindent
\textbf{Learning Latent Confounder.}
We use  iVAE \cite{khemakhem2020variational} to learn the latent confounder, 
since it is widely used to identify latent variables up to an equivalence relation (see the Definition 2 in  \cite{khemakhem2020variational})
by leveraging auxiliary variables which are equivalent to proxies.
Specifically, we simultaneously learn the deep generative model and approximate posterior $q_\phi(\adcub|\au,\wu)$ of the true posterior $p_\theta(\adcub|\au,\wu)$
by maximizing $\mathcal{L}(\theta,\phi)$, which is the evidence lower bound (ELBO) of the likelihood $\log p_\theta(\au|\wu)$:
\begin{equation}
\label{eq: ELBO}
\begin{split}
     &   E[\log p_\theta( \au | \wu)] \geq \mathcal{L} (\theta,\phi):= \\ 
      =&  E[
      \underbrace{E_{q_\phi(\adcub | \au, \wu)}
      [ \log p_\theta(\adcub|\wu) -\log \qphi]}_{\uppercase\expandafter{\romannumeral1}} \\  
       &+ \underbrace{E_{q_\phi(\adcub | \au, \wu)}
      [ \log p_\theta(\au|\adcub)]}_{ \uppercase\expandafter{\romannumeral2}}
].
\end{split}
\end{equation}
where according to the causal graph in \cref{fig: proximal causal graph}, 
$\log  p_\theta(\au,\adcub|\wu)$ is further decomposed as follows:
\begin{equation}
\begin{split}
         \log  p_\theta(\au,\adcub|\wu)
        =&\log p_\theta(\au|\adcub,\wu)+ \log p_\theta(\adcub|\wu) \\
        =&\log p_\theta(\au|\adcub)+\log p_\theta(\adcub|\wu).
\end{split}
\end{equation}

Following \cite{khemakhem2020variational}, we choose the prior $p_\theta (\adcub|\wu)$  to be a Gaussian location-scale family,
and use the reparameterization trick \cite{kingma2013auto} to sample 
$\adcu$ from the approximate posterior $q_\phi(\adcub|\au,\wu)$ as
\begin{equation}
\begin{split}
    p_\theta(\adcub|\wu) &:=N(\mu_w(\wu),v_w(\wu)), \\
    q_\phi(\adcub|\au,\wu)&:=N(\mu_{aw}(\au,\wu), v_{aw}(\au,\wu)),
\end{split}
\end{equation}
where $\mu_w,v_w,\mu_{aw},v_{aw}$ are modeled by $4$ different MLP models.
To this end, the calculation of the expectation 
$\uppercase\expandafter{\romannumeral1}$ of Eq. (11)
can be converted to the calculation of the Kullback-Leibler divergence of two Gaussian distributions:
\begin{equation}
\nonumber
\begin{split}
        &E_{q_\phi(\adcub|\au,\wu)}[\log p_\theta(\adcub|\wu)-\log \qphi]\\
        =&-KL(N(\mu_{aw}(\au,\wu), v_{aw}(\au,\wu), N(\mu_w(\wu),v_w(\wu))).        
\end{split}
\end{equation}

As for $\uppercase\expandafter{\romannumeral2}$, since the hidden confounder directly affects each element of the exposure vector,
we use a factorized logistic model as $p_\lambda(\au|\zu)$, i.e.,
$p_\lambda(\au|\zu)=\prod_{i=1}^n Bernoulli(\boldsymbol{a}_{ui}|\zu)$,
which is also modeled by a MLP $\mu_z(\vz)$.
Then the log-likelihood $\log p_\theta(\au|\adcub)$ becomes the negative binary cross entropy:
\begin{equation}
\nonumber
        \log p_\theta(\au|\adcub)= \sum_{i=1}^n \va_{ui} \log (\mu_z(\zu)_i) + (1-\va_{ui})\log (1-\mu_z(\zu)_i).
\end{equation} 

Then, through maximizing \cref{eq: ELBO}, we are able to obtain the approximate posterior of latent confounder $\qphi$.  


\noindent\textbf{Feedback with given latent confounder.}
As shown in \cref{eq: factor model with w},  with $q_\phi(\adcub|\au,\wu)$ estimated through iVAE, the user's feedback on item $i$ 
with the latent confounder $\adcub$, i.e., $p(\Rui|\au, \adcub)$, can be learned by fitting a recommendation model on the observed users' feedback.
Following the assumption in \cref{sec: 3.1} where 
$\Ruia$ is only affected by the exposure of item $i$ to user $u$,
we use a point-wise recommendation model $f(u,i,\vz_u; \eta)$ parameterized by $\eta$ to estimate $p(\Rui|\au, \adcub)$.
Specifically, we adopt a simple additive model  
$f(u,i,\adcub; \eta) = f_1(u,i)+f_2(\adcub, i)$ 
that models the user's intrinsic preference and
the effect of the latent confounder separately. 
The corresponding loss function is:
\begin{equation}
\label{eq: rating loss abstract}
\mathcal{L}_{\modelname} (\eta) = \frac{1}{|\mathcal{D}|}
\sum_{(u,i)\in \mathcal{D}} 
l(E_{q_\phi (\adcub |\au, \wu)}[f(u,i,\adcub; \eta)] , \Rui), 
\end{equation}
where $l(\cdot,\cdot)$ is one of the commonly-used loss functions for recommendation systems, e.g., MSE loss and BCE loss.


\noindent
\textbf{Inference Stage.}
In practice, for most real-world recommendation datasets,
the user's feature $\wu$ is invariant in the training set and test set.
Therefore, identifying $p(\Ruia)$ is equivalent to identifying $p(\Ruia|\wu)$ since $p(\Ruia)=\int_w p(\Ruia|\wu=w)p(\wu=w) dw$
and $p(\wu=w)=1$ for those specific $w$ associated with user $u$.
The corresponding identification formula becomes:
\begin{equation}
\begin{split}
\label{eq: adjust the prediction}
     p(\Ruia|\wu) &=  \int_{\vz} p(\adcub=\vz|\wu)p(\Rui|\va,\adcub=\vz) d \vz \\
              &=E_{\adcub|\wu}[p(\Rui|\va,\adcub)],
\end{split}
\end{equation}
where $p(\Rui|\va,\adcub= \vz)$ is estimated 
by the learned recommendation model $f(u,i,\vz_u; \eta)$ 
and 
$p(\adcub=\vz|\wu)$ is approximated by the encoder $q_\phi(\adcub|\au,\wu)$.


In summary, 
we first apply iVAE to learn the posterior distribution of the  latent confounder $p(\adcu|\au,\wu)$ for each user $u$,
then leverage it to learn the user's feedback estimator 
$f(u,i,\vz_u;\eta)$ 
in the training phase.
Finally, we apply \cref{eq: adjust the prediction} to predict the deconfounded feedback in the inference phase.
The pseudo-code of \modelname is shown in Algorithm \ref{alg}.

\section{Experiments}
In this section, we conduct experiments to answer the following research questions:
\begin{itemize}[noitemsep,topsep=0pt,leftmargin=*]
    \item \textbf{RQ1} Does the proposed  \modelname outperform existing deconfounding methods for debiasing recommendation systems? 
    \item \textbf{RQ2} What is the performance of \modelname under different confounding effects and dense ratios of the exposure matrix?

    \item  \textbf{RQ3} How does the identification of latent confounders impact the performance of iDCF?
\end{itemize} 

\subsection{Experiment Settings}

\noindent\textbf{Dataset.} Following previous work \cite{address,deconfounder,wang2022invariant}, we perform experiments on three real-world datasets: Coat \footnote{https://www.cs.cornell.edu/~schnabts/mnar/},
Yahoo!R3 \footnote{https://webscope.sandbox.yahoo.com/}
and KuaiRand\footnote{https://kuairand.com/} collected from different recommendation scenarios. 
Each dataset consists of a biased dataset of normal user interactions, 
and an unbiased uniform dataset collected by a randomized trial such that users will interact with randomly selected items.
We use all biased data as the training set, 
    30\% of the unbiased data as the validation set,
    and the remaining unbiased data as the test set.
    For Coat and Yahoo!R3,
    the feedback from a user to an item is a rating ranging from 1 to 5 stars.
    We take the ratings  $\geq 4$ as positive feedback, and others as negative feedback.
    For KuaiRand, the positive samples are defined according to the signal "IsClick" provided by the platform.

Moreover, to answer RQ2 and RQ3,  we also generate a synthetic dataset with groundtruth of the unmeasured confounder known
for in-depth analysis of the \modelname.


\begin{table}
\caption{The statistic of Coat, Yahoo!R3, and KuaiRand.}

    \centering
\begin{tabular}{ c c c c c  } 
 \hline
 Dataset & \#User & \#Item & \#Biased Data & \#Unbiased Data\\ 
 \hline
 Coat & 290 & 300 & 6,960 & 4,640\\ 
 Yahoo! R3 & 5,400 & 1,000 & 129,179 & 54,000\\ 
 KuaiRand  & 23,533 & 6,712 & 1,413,574 & 954,814 \\ 
 \hline
\end{tabular}
\label{table: statistic}
\vspace{-4mm}
\end{table}

\noindent\textbf{Baselines.} 
We compare our method \footnote{https://github.com/BgmLover/iDCF} with the corresponding base models and the state-of-the-art deconfounding methods that can alleviate the confounding bias in recommendation systems in the presence of unmeasured confounders.

\begin{itemize}[noitemsep,topsep=0pt,leftmargin=*]
    
    \item MF \cite{koren2009matrix} \& MF with feature (MF-WF). 
    We use the classical Matrix Factorization (MF) as the base recommendation model.
    Since our method utilizes user features, for a fair comparison, we consider MF-WF, a variant of MF model augmented with user features.

    \item DCF \cite{deconfounder}.
    Deconfounder (DCF) addresses the unmeasured confounder by learning a substitute confounder to approximate the true unmeasured confounder and applying the g-formula for debiasing. 
    However, as discussed before, it fails to guarantee the identification of users' feedback, leading to the inconsistent prediction of users' feedback. 
    
    
    \item IPS \cite{IPS-16} \& RD-IPS \cite{address}.
    IPS is a classical propensity-based deconfounding method that ignores the unmeasured confounder and directly leverages the exposure to estimate propensity scores to reweight the loss function.
    RD-IPS is a recent IPS-based deconfounding method that assumes the bounded confounding effect of the unmeasured confounders to derive bounds of propensity scores 
    and applies robust optimization for robust debiasing. 
    The implementation of the two methods leverages a small proportion of unbiased data to get more accurate propensity scores.
    
    
    \item InvPref \cite{wang2022invariant}. InvPref assumes the existence of multiple environments as proxies of unmeasured confounders and applies invariant learning \cite{buhlmann2020invariance,arjovsky2019invariant} to learn the user's invariant preference.
    \item DeepDCF-MF.
    DeepDCF~\cite{zhu2022deep} extends DCF by applying deep models
    and integrating the user's feature into the feedback prediction model to control the variance of the model.
    For a fair comparison, we adapt their model with  MF as the backbone model.
    
    \item iDCF-W.
    iDCF-W is a variant of iDCF that does not leverage proxy variables.
    We adopt VAE \cite{kingma2013auto} to learn the substitute confounder in such a scenario, with other parts staying the same with iDCF.  
\end{itemize}

\noindent
\textbf{Implementation details.}  Due to space limitations, please refer to \cref{appendix: exp detail}.

\subsection{Performance Comparison (RQ1)}
The experimental results on the three real-world datasets are shown in \cref{table: real-world dataset}. We can observe that: 
\begin{table*}[!h]
\caption{Recommendation performances on Coat, Yahoo!R3 and KuaiRand. The p-value under t-test between iDCF and the best baseline on each dataset is also provided. 
}
\centering
\begin{tabular}{c|ll|ll|ll} 
\hline
\multicolumn{1}{l|}{\multirow{2}{*}{Datasets}} & \multicolumn{2}{c|}{Coat}                              & \multicolumn{2}{c|}{Yahoo!R3}                                  & \multicolumn{2}{c}{KuaiRand}                                \\
\multicolumn{1}{l|}{}                          & \multicolumn{1}{c}{NDCG@5} & \multicolumn{1}{c|}{RECALL@5} & \multicolumn{1}{c}{NDCG@5} & \multicolumn{1}{c|}{RECALL@5} & \multicolumn{1}{c}{NDCG@5} & \multicolumn{1}{c}{RECALL@5}  \\ 
\hline
MF & $0.5524 \pm  0.0144$ & $0.5294 \pm  0.0227$ & $0.5629 \pm  0.0100$ & $0.7129 \pm  0.0106$ & $0.3748 \pm  0.0018$ & $0.3247 \pm  0.0013$ \\
MF-WF & $\underline{0.5529} \pm  0.0101$ & $\underline{0.5341} \pm  0.0143$ & $0.5649 \pm  0.0073$ & $0.7144 \pm  0.0086$ & $0.3762 \pm  0.0014$ & $0.3255 \pm  0.0013$ \\
IPS & $0.5450 \pm  0.0161$ & $0.5260 \pm  0.0191$ & $0.5490 \pm  0.0058$ & $0.6967 \pm  0.0096$ & $0.3696 \pm  0.0011$ & $0.3224 \pm  0.0009$ \\
RD-IPS & $0.5448 \pm  0.0147$ & $0.5240 \pm  0.0157$ & $0.5550 \pm  0.0051$ & $0.7020 \pm  0.0068$ & $0.3690 \pm  0.0016$ & $0.3207 \pm  0.0011$ \\
InvPref & $0.5405 \pm  0.0135$ & $0.5295 \pm  0.0225$ & $0.5928 \pm  0.0038$ & $0.7414 \pm  0.0052$ & $0.3778 \pm  0.0020$ & $0.3283 \pm  0.0014$ \\
DCF & $0.5509 \pm  0.0093$ & $0.5329 \pm  0.0152$ & $0.5675 \pm  0.0047$ & $0.7116 \pm  0.0059$ & $0.3751 \pm  0.0015$ & $0.3243 \pm  0.0012$ \\
DeepDCF-MF & $0.5373 \pm  0.0066$ & $0.5141 \pm  0.0113$ & $0.6395 \pm  0.0044$ & $0.7729 \pm  0.0056$ & $\underline{0.4078} \pm  0.0013$ & $\underline{0.3491} \pm  0.0010$ \\
iDCF-W & $0.5255 \pm  0.0137$ & $0.4971 \pm  0.0183$ 
& $\underline{0.6410} \pm  0.0029$ & $\underline{0.7712} \pm  0.0033$ 
& $0.4072 \pm  0.0009$ & $0.3481 \pm  0.0011$ \\
iDCF (ours) &  $\textbf{0.5744} \pm  0.0122$ & $\textbf{0.5504} \pm  0.0126$ & $\textbf{0.6455} \pm  0.0023$ & $\textbf{0.7837} \pm  0.0035$ 
& $\textbf{0.4093} \pm  0.0004$ & $\textbf{0.3513} \pm  0.0009$ \\
\hline
p-value & $7e^{-4}$ & $2e^{-2}$ & $2e^{-3}$ & $1e^{-4}$ & $5e^{-3}$ & $1e^{-4}$ \\
\hline
\end{tabular}

\label{table: real-world dataset}
\vspace{-0mm}
\end{table*}

\begin{itemize} [noitemsep,topsep=0pt,leftmargin=*]
\item
    The proposed \modelname consistently outperforms the baselines with statistical significance suggested by low p-values w.r.t. all the metrics across all datasets,
    showing the gain in empirical performance due to the identifiability of counterfactual feedback by inferring identifiable latent confounders. This is further verified by experimental results in the synthetic dataset (see \cref{sec: simulation}).    
        
    \item DCF, DeepDCF-MF, iDCF-W and iDCF achieve better performance than the base models (MF and MF-WF) in Yahoo!R3 and KuaiRand. 
    This implies that leveraging the inferred hidden confounders to predict user preference can improve the model performance when the sample size is large enough.
    Moreover, deep latent variable models (VAE, iVAE) perform better than the simple Poisson factor model in learning the hidden confounder with their ability to capture nonlinear relationships between the treatments and hidden confounders. 

    \item However, the poor performance of DeepDCF-MF, iDCF-W, and DCF in Coat shows the importance of the identification of the feedback through learning the identifiable latent confounders. While the proposed \modelname provides the guarantee on the identification of the counterfactual feedback in general,
    these methods cannot guarantee the identification of the feedback.
    Therefore, iDCF outperforms DeepDCF-MF in all cases, 
    even though they take the same input and use similar MF-based models for feedback prediction.

    \item MF-WF slightly outperforms MF in all cases, 
    showing that incorporating user features into the feedback prediction model improves the performance. 
    Moreover, DeepDCF-MF outperforms iDCF-W in all datasets except Yahoo!R3.
    Note that DeepDCF-MF incorporates user features into the feedback prediction model while iDCF-W does not.
    This implies that the effectiveness of incorporating user features into feedback prediction depends on whether the user features are predictive of the user preference.
    For example, in Yahoo!R3, the user features are from a questionnaire that contains questions about users' willingness to rate different songs that might influence their exposure but are not directly related to their feedback. 
    DeepDCF-MF directly incorporates such user features into the feedback prediction model,
    which introduces useless noise.
    This may explain why DeepDCF-MF is outperformed by iDCF-W in this dataset. 
    
\end{itemize}

\subsection{In-depth Analysis with Synthetic Data  (RQ2 \& RQ3)}
\label{sec: simulation}

Our method relies on the inference of the unmeasured confounder. However, in real-world datasets, the ground truth of unmeasured confounders is inaccessible.
To study the influence of learning identifiable latent confounders on the recommendation performance, we create a synthetic dataset (see \cref{appendix: exp detail} for details) to provide the ground truth of the unmeasured confounder.

There are three important hyper-parameters in the data generation process:
$\alpha$ controls the density of the exposure vector, a larger $\alpha$ means a denser exposure vector.
    $\beta$ is the weight of the confounding effect of the user's preference,
    a larger $\beta$ means the confounder has a stronger effect on the user's feedback.
    $\gamma$ controls the weight of the random noise in the user's exposure,
    a larger $\gamma$ means the user's exposure is more random.
Similar to the real-world datasets, 
for each user, 
we randomly select $15$ items and collect these data as the unbiased dataset.
The data pre-processing is the same as the experiments on real-world dataset in Section 5.1-5.2.

\noindent \textbf{RQ2: Performance of \modelname under different confounding effects and dense ratio of the exposure matrix.}
We conduct experiments on the simulated data to study the robustness of our method.
The results show that \modelname is robust and can still perform well under varying confounding effects and dense ratios. 

\noindent
\textbf{Effect of confounding weight.}
We fix the dense ratio $\alpha=0.1$ and the exposure noise weight $\gamma=0$ , then vary the confounding weight $\beta$.
Recall a larger $\beta$ means a stronger confounding effect.

%
The result is shown in \cref{table: cr} and we find that:
\begin{itemize}[noitemsep,topsep=0pt,leftmargin=*]
    \item The proposed method \modelname  outperforms the baselines in all cases with small standard deviations.
    \item As the confounding effect $\beta$ increases, the performance gap between \modelname and the best baselines becomes more significant, measured by both the mean NDCG@5, Recall@5 and the p-value.
    This justifies the effectiveness of deconfounding of \modelname.
\end{itemize}

\begin{table*}
\caption{Recommendation performances on the simulated datasets with different confounding effects. 
A larger $\beta$ results in a stronger confounding effect.
The p-value under t-test between iDCF and the best baseline is also reported.
}
\vspace{-0mm}
\centering
\begin{tabular}{c|ll|ll|ll} 
\hline
\multicolumn{1}{l|}{\multirow{2}{*}{Datasets}} & \multicolumn{2}{c|}{$\beta=1.0$}                              & \multicolumn{2}{c|}{$\beta=2.0$}                                  & \multicolumn{2}{c}{$\beta=3.0$}                                \\
\multicolumn{1}{l|}{}                          & \multicolumn{1}{c}{NDCG@5} & \multicolumn{1}{c|}{RECALL@5} & \multicolumn{1}{c}{NDCG@5} & \multicolumn{1}{c|}{RECALL@5} & \multicolumn{1}{c}{NDCG@5} & \multicolumn{1}{c}{RECALL@5}  \\ 
\hline
MF & $0.7911 \pm  0.0022$ & $0.6724 \pm  0.0020$ & $0.8029 \pm  0.0018$ & $0.6593 \pm  0.0018$ & $0.8217 \pm  0.0026$ & $\underline{0.6423} \pm  0.0023$ \\
MF-WF & $0.7914 \pm  0.0021$ & $0.6716 \pm  0.0020$ & $0.8028 \pm  0.0023$ & $0.6588 \pm  0.0019$ & $0.8220 \pm  0.0030$ & $0.6414 \pm  0.0029$ \\
DCF & $0.7904 \pm  0.0019$ & $0.6720 \pm  0.0013$ & $0.8024 \pm  0.0025$ & $0.6593 \pm  0.0029$ & $0.8223 \pm  0.0031$ & $0.6423 \pm  0.0035$ \\
IPS & $0.7890 \pm  0.0026$ & $0.6706 \pm  0.0017$ & $0.7982 \pm  0.0014$ & $0.6552 \pm  0.0020$ & $0.8159 \pm  0.0038$ & $0.6379 \pm  0.0028$ \\
RD-IPS & $0.7878 \pm  0.0058$ & $0.6694 \pm  0.0029$ & $0.8001 \pm  0.0022$ & $0.6569 \pm  0.0027$ & $0.8193 \pm  0.0026$ & $0.6396 \pm  0.0021$ \\
InvPref & $\underline{0.7953} \pm  0.0027$ & $\textbf{0.6761} \pm  0.0033$ & $0.7985 \pm  0.0029$ & $0.6556 \pm  0.0036$ & $0.8144 \pm  0.0051$ & $0.6358 \pm  0.0039$ \\
DeepDCF-MF & $0.7917 \pm  0.0017$ & $0.6715 \pm  0.0019$ & $\underline{0.8060} \pm  0.0032$ & $\underline{0.6601} \pm  0.0029$ & $0.8220 \pm  0.0026$ & $0.6421 \pm  0.0028$ \\
iDCF-W & $0.7901 \pm  0.0010$ & $0.6703 \pm  0.0015$ 
& $0.8050 \pm  0.0029$ & $0.6590 \pm  0.0040$ 
& $\underline{0.8226} \pm  0.0015$ & $0.6420 \pm  0.0008$ \\
iDCF (ours) & $\textbf{0.7973} \pm  0.0023$ & $\underline{0.6735} \pm  0.0020$ & $\textbf{0.8168} \pm  0.0013$ & $\textbf{0.6683} \pm  0.0015$ & $\textbf{0.8368} \pm  0.0019$ & $\textbf{0.6549} \pm  0.0025$ \\
\hline
p-value & $1e^{-1}$ & $6e^{-2}$ & $2e^{-8}$ & $6e^{-7}$ & $8e^{-13}$ & $1e^{-9}$ \\
\hline
\end{tabular}

\label{table: cr}
\vspace{-3mm}
\end{table*}

\begin{figure}[h]
\center
\vspace{-2mm}
\begin{subfigure}{0.23 \textwidth}
\includegraphics[width=1\textwidth]{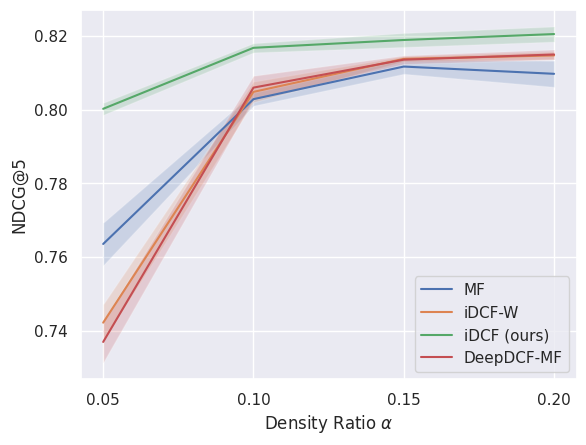}
\caption{}   
\label{fig: sr}
\end{subfigure}
\begin{subfigure}{0.23\textwidth}
\includegraphics[width=1\textwidth]{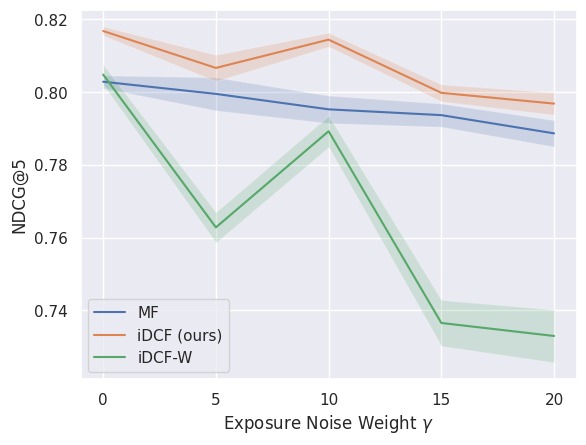}
\caption{}
\label{fig: tr}
\end{subfigure}
\vspace{-4mm}
\caption{
Recommendation performance on the simulated datasets with different (a) exposure density ratios  and (b) exposure noise weights.
A larger $\alpha$ means denser user exposure.
A larger $\gamma$ means the exposure contains more random noise.
} 
\label{fig: tr_2}


\center
\begin{subfigure}{0.155 \textwidth}
\includegraphics[width=1.1\textwidth]{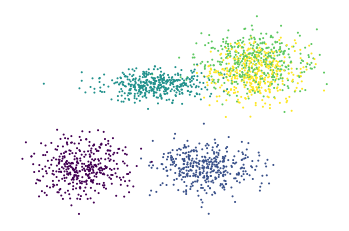}
\caption{Ground truth.}   
\label{fig: true confounder}
\end{subfigure}
\begin{subfigure}{0.155\textwidth}
\includegraphics[width=1.1\textwidth]{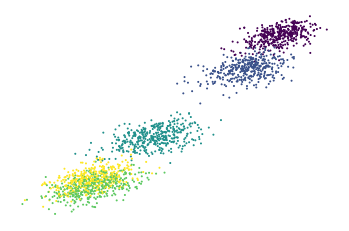}
\caption{\modelname (ours).}
\label{fig: ivae}
\end{subfigure}
\begin{subfigure}{0.155\textwidth}
\includegraphics[width=1.1\textwidth]{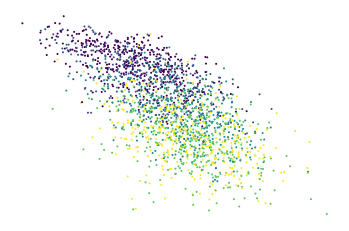}
\caption{iDCF-W.}
\label{fig: vae}
\end{subfigure}
\vspace{-6mm}
\caption{The visualization of the true unmeasured confounder and the learned latent confounders using \modelname and iDCF-W.
The colors correspond to the user's feature $\wu$.
} 
\label{fig: visual confounders}
\vspace{-4mm}
\end{figure}

\noindent
\textbf{Effect of density of exposure vector.}
Next, we investigate the performance  of \modelname under different dense ratios $\alpha$ by fixing $\beta=2.0, \gamma=0$. 
Due to space limitations, we only report NDCG@5 of the best four methods in \cref{table: cr} in  \cref{fig: sr}.
%
It can be found that:
\begin{itemize}[noitemsep,topsep=0pt,leftmargin=*]
    \item Overall, all the recommendation methods achieve better performances with less sparse data as $\alpha$ increases.
\item Similar to the observations in the Coat dataset, \modelname is more robust than the baselines when exposure becomes highly sparse. At the same time, iDCF-W and DeepDCF-MF achieve very poor performance with highly sparse data with small $\alpha$. This further verifies the efficacy of learning identifiable latent confounders.
    
\end{itemize}


\noindent \textbf{RQ3: Influence of learning identifiable latent confounders.}
The synthetic dataset enables us to visualize the true unmeasured confounder and study the influence of the identifiability of the learned confounders on the model performance.
Here, we show the identification of the latent confounder by visualization, 
and conduct experiments to study the robustness of \modelname against different exposure noise weights $\gamma$ with fixed $\alpha=0.1$ and $\beta=2.0$.
The empirical results show that our method can better identify the unmeasured confounder, 
leading to more accurate feedback predictions.

\noindent \textbf{Visualization of the learned latent confounder.}
\cref{fig: true confounder} shows the conditional distributions of the two-dimensional \emph{ground truth} of the unmeasured confounder $P(\zu|\wu)$ with the exposure noise weight $\gamma=0$. 
We use \modelname and iDCF-W  to learn the corresponding latent confounders, respectively, 
and we plot the posterior distributions $p(\adcub|\au, \wu)$ and $p(\adcub|\au)$ in \cref{fig: ivae,fig: vae}.
It can be shown that \modelname can identify a better latent confounder than iDCF-W does, which helps to explain the observation that \modelname is better than iDCF-W in previous experiments.

\noindent \textbf{Impact of the exposure noise on the learned confounder.}
Next, we vary the exposure noise weight $\gamma$ to study the impact of the $\gamma$ on the learned latent confounder.
The intuition behind this experiment is that as the weight of the noise increases, there will be more randomness in the exposure vectors, making it more challenging to infer the unmeasured confounders.



To assess the accuracy of the learned confounders in approximating the ground truth, we compute the mean correlation coefficients (MCC) between the learned confounders and the ground truth. MCC is a widely accepted metric in the literature for evaluating the identifiability of learned latent variables \cite{khemakhem2020variational}. The results are presented in \cref{table: tr}. As observed in the table, the results suggest that as the noise level increases, it becomes increasingly difficult to approximate the ground truth using the learned confounders,
while iDCF is much more robust regarding to the increasing noise level, compared to iDCF-W.

\begin{table}
\caption{The mean correlation coefficients (MCC) between the true confounder and the latent confounders learned by \modelname and iDCF-W model.
A larger MCC means a larger correlation with the true unmeasured confounder.}
\centering

\begin{tabular}{c|l|l|l|l|l} 
\hline
Model & $\gamma=0.$ & $\gamma=5.0$ & $\gamma=10.0$ & $\gamma=15.0$ 
& $\gamma=20.0$ 
\\
\hline
iDCF-W & $0.6050$ & $0.3374$ & $0.1023$ & $0.1001$ 
& $0.0682$ 
\\
iDCF (ours) & $0.8394$ & $0.8052$ & $0.6955$ & $0.6914$ 
& $0.6062$ 
\\
\hline
\end{tabular}

\label{table: tr}
\vspace{-6mm}
\end{table}

\noindent \textbf{Impact of exposure noise on the feedback prediction.}
Moreover, we conduct experiments to investigate how the performance of \modelname varies with the exposure noise weight $\gamma$.
We choose MF and iDCF-W as the baselines 
because (1) MF is an empirically stable recommendation model and (2) iDCF-W is the same as \modelname except it does not guarantee the identifiability of the learned confounders.
We report NDCG@5 in \cref{fig: tr}.
The results indicate that, in general, as the exposure noise increases, it becomes more challenging to identify latent confounders, which in turn makes it more difficult to predict counterfactual feedback.
These results, along with those in \cref{table: tr}, 
show that a better approximation of the ground truth confounders often leads to better estimation of the true user feedback.





\section{conclusion and future work}
In this work, 
we studied how to identify the user's counterfactual feedback by mitigating the unmeasured confounding bias in recommendation systems.
We highlight the importance of identification of the user's counterfactual feedback
by showing the non-identification issue of the Deconfounder method, which can finally lead to inconsistent feedback prediction.
To this end,
we propose a general recommendation framework that utilizes proximal causal inference to address the non-identification issue 
and provide theoretical guarantees for mitigating the bias caused by unmeasured confounders.
We conduct extensive experiments to show the effectiveness and robustness of our methods in real-world datasets and synthetic datasets.

This work leverages proxy variables to infer the unmeasured confounder and users' feedback.
In the future, we are interested in trying more feasible proxy variables (e.g., item features)
and how to combine different proxy variables to achieve better performance. 
It also makes sense to apply our framework to sequential recommendations and other downstream  recommendation scenarios (e.g., solving the challenge of filter bubbles).
\bibliographystyle{ACM-Reference-Format}
\balance
\bibliography{ref}

\appendix

\section{Algorithm}
\begin{algorithm}
    \SetKwInOut{Input}{Input}\SetKwInOut{Output}{Output}
    \caption{Identifiable Deconfounder (iDCF)}
    \label{alg}
    \Input{$\{\va_u, \vw_u\}, \forall u \in \mathcal{U}$, $\{r_{ui}\}, \forall (u,i) \in \mathcal{D}$ }
    \textbf{Training phase:}\\
    \tcp{{Learning Latent Confounder}}
    
    Calculate the latent confounder distribution $q_\phi(\adcub|\au,\vw_u)$ for each user $u$ by maximizing \cref{eq: ELBO}\;
    
   \tcp{Feedback with given latent confounder}
   
    Initialize a recommendation model $f(u, i,\adcub; \eta)$ with parameters $\eta$\;
    \While{Stop condition is not reached}
    {
    Fetch $(u,i)$ from $\mathcal{D}$\;
    Minimize the loss \cref{eq: rating loss abstract} to optimize $\eta$;
    }
    \textbf{Inference phase:}\\
    Calculate the prediction of user's feedback $\hat{r}_{ui}$ according to \cref{eq: adjust the prediction} for each $(u,i)$ pair.
\end{algorithm}

\section{Experiment Details}
\label{appendix: exp detail}
\noindent\textbf{Data Generation Process.}
The simulated dataset consists of 2,000 users and 300 items. 
For each user $u$, 
the unmeasured confounder $\zu$ is a two-dimensional representation of the user's socio-economic status sampled from a mixture of five independent multivariate Gaussian distributions.
The proxy variable $w_u \in \{1, 2, 3, 4, 5\}$ is a one-dimensional categorical variable indicating the user's consumption level, which is determined by $\zu$ such that the prior of $w_u$ is uniformly distributed and the conditional distribution of $\zu$ follows: 
\begin{equation}
\zu^k|w_u \sim N(\mu_k(w_u), \sigma_k^2(w_u)), k \in \{1,2\},
\end{equation}
where ${\zu}^k$ is the j-th element of $\zu$.

%
The exposure of the pair $(u,i)$ is generated by
\begin{equation}
    \begin{split}
        & a_{ui} \sim Bernoulli(g_i(\zu)), \\
& g_i(\vz)=\alpha \cdot sigmoid(LeakyRelu(\vz \boldsymbol{M}  e_{zi}) + \gamma \epsilon)  
    \end{split}
\end{equation}
where $M$ is a $2 \times 2$ matrix,
and each element of $M$ is sampled from a uniform distribution.
$e_{zi}$  is a randomly generated item-wise 2-dimensional  embedding vector, 
$\alpha$ is a hyper-parameter that controls the sparsity of the exposure vector $\au$,
$\epsilon$ is random noise,  
and $\gamma$ is the corresponding weight of the noise. 

The true feedback of user $u$ on item $i$ is $\Rui= f_n(e_u ^T e_i + \beta z_u^T e_{zi}  + \epsilon_{ui})$,
where $f_n: \mathbb{R} \to \{1,2,3,4,5\}$ is a normalization function, 
$\epsilon_{ui}$ is an i.i.d. random noise,
and $\beta$ is a hyper-parameter controlling the weight of the confounding effect.

\noindent\textbf{Implementation Details.} 

\noindent\textit{Outcome Model.} Our method is model-agnostic in the sense that it works with any outcome prediction model. For ease of comparison,
we follow the recent work on unmeasured confounders~\cite{deconfounder}, and 
adopt matrix factorization (MF)  as the backbone model.
Specifically, we take $f(u,i,\adcub; \eta) = f_1(u,i)+f_2(\adcub,i)$ in \cref{eq: rating loss abstract}, where 
\begin{equation}
    f_1(u,i) = \embe_u^T \embe_i+ b_u + b_i,
    f_2(\adcub,i) = \adcub^T \embc_i,
\end{equation}
where $\embe_i,\embc_i$ are different embeddings of item $i$, $\embe_u$ is embedding representation of user $u$,
$b_u,b_i$ are the user preference bias term and item preference bias term, respectively.
During training, $\adcub$ is sampled from $\qphi$ to approximate the integral in \cref{eq: factor model with w}.

In the inference phase, 
we direct take $\bar{\vz}_u=E_{q_\phi(\adcub|\au,\wu)}[\adcub]$ and estimate the user's feedback on item $i$ as follows:
\begin{equation}
        \hat{r}_{ui}= \embe_u^T \embe_i + \bar{\vz}_u^T \embc_i + b_u + b_i.
\end{equation} 
%

\noindent\textit{Hyper-parameter search.} 
For all recommendation models, we use grid search to select the hyper-parameters based
on the model’s performance on the validation dataset.
The learning rate is searched from \{1e-3, 5e-4, 1e-4, 5e-5, 1e-5\}, and the weight decay is chosen from \{1e-5, 1e-6\}. 
We adopt their codes for the baselines with public implementation and follow the suggested range of hyper-parameters.
The public implementation of IPS and RD-IPS \cite{address} relies on a small set of unbiased data to obtain the propensity scores, 
we follow their procedure and extract the same proportion of unbiased data from the validation set.
For a fair comparison, we use ADAM \cite{kingma2014adam} for the optimization of all models.


\noindent\textbf{Evaluation Metrics} are $NDCG@K$ and $Recall@K$.
We report the average value and standard deviation for each method with $10$ different random seeds.
The p-value of the T-test between our method and the best baseline is also reported.

\section{Supplementary Proof}
\begin{proof}[Proof of \cref{lemma: identification in example}]

There are 4 unknown values $\{p_{zr|a}, z,r \in\{0,1\}\}$ with 4 linear constraints: 
\begin{equation}
\label{eq: equations}
\begin{array}{ll}
       (1) \sum_{z } \sum_{r} p_{zr|\va}=1, & 
        (2) \sum_{z} p_{z1|\va} \frac{\pi_{\adcu=z|\va ,w=1}}{\pi_{\adcu=z|\va}}=\pi_{\Rui=1|\va,w=1}, \\
        (3) \sum_{r} p_{1r|\va}=\pi_{\adcu=1|\va},  
        & 
        (4) \sum_{z} p_{z1|a}  \frac{\pi_{\adcu=z|\va,w=0}}{\pi_{\adcu=z|\va}}=\pi_{\Rui=1|\va,w=0}.
\end{array}
\end{equation}

By solving (1) and (3):
\begin{equation}
    p_{10} = \pi_{\adcu=1|\va} - p_{11}, p_{00}= 1 - p_{01} - \pi_{\adcu=1|\va}.
\end{equation}
Then it remains to solve $p_{01}$ and $p_{11}$ from (2) and (4).
and the unique solution from (2) and (4) requires that
\begin{equation}
\label{eq: unique condition}
\begin{split}
    \frac{\pi_{\adcu=0|\va,w=1}}{\pi_{\adcu=0|\va}} \frac{\pi_{\adcu=1|\va,w=0}}{\pi_{\adcu=1|\va}}
     -  \frac{\pi_{\adcu=1|\va,w=1}}{\pi_{\adcu=1|\va}} \frac{\pi_{\adcu=0|\va,w=0}}{\pi_{\adcu=0|\va}} & \neq 0 \\ 
     \pi_{\adcu=1|\va,w=1}(1-\pi_{\adcu=1|\va,w=0}) 
     - \pi_{\adcu=1|\va,w=0}(1-\pi_{\adcu=0|\va,w=1}) & \neq 0 \\
     \pi_{\adcu=1|\va,w=1} 
     - \pi_{\adcu=1|\va,w=0} & \neq 0. 
\end{split}
\end{equation}

Since $w_u$ is assumed to be correlated with $z_u$ condition on $\va$,
and $\adcu$ is learned from the unique factorization $p(\au)$,
which means $\adcu$ is also correlated with $w_u$ condition on $\va$.

Therefore, the condition in \cref{eq: unique condition} is satisfied,
i.e., $\{p_{zr|a}, z,r \in\{0,1\}\}$ have a unique solution.
And $\Ruia$ is also uniquely determined by \cref{eq: example_identification}.

\end{proof}
\end{document}